\documentclass[aip,
 amsmath,amssymb,
 reprint,%
]{revtex4-1}

\usepackage{graphicx}
\usepackage{dcolumn}
\usepackage{bm}

\usepackage[utf8]{inputenc}
\usepackage[T1]{fontenc}
\usepackage{mathptmx}
\usepackage{etoolbox}

\makeatletter
\def\@email#1#2{%
 \endgroup
 \patchcmd{\titleblock@produce}
  {\frontmatter@RRAPformat}
  {\frontmatter@RRAPformat{\produce@RRAP{*#1\href{mailto:#2}{#2}}}\frontmatter@RRAPformat}
  {}{}
}%
\makeatother
\begin{document}

\preprint{AIP/123-QED}

\title[Benchmarking LLMs for materials synthesis: atomic layer
deposition]{Benchmarking large language models for materials synthesis: the case of atomic layer deposition}
\author{Angel Yanguas-Gil}
 \email{ayg@anl.gov}
\affiliation{Applied Materials Division, Argonne National Laboratory, Lemont, IL 60439 USA}
\author{Matthew T. Dearing}
\affiliation{Business and Information Systems, Argonne National Laboratory, Lemont, IL 60439 USA}
\author{Jeffrey W. Elam}
\author{Jessica C. Jones}
\author{Sungjoon Kim}
\author{Adnan Mohammad}
\author{Chi Thang Nguyen}
\affiliation{Applied Materials Division, Argonne National 
Laboratory, Lemont, IL 60439 USA}
\author{Bratin Sengupta}
\affiliation{Applied Materials Division, Argonne National 
Laboratory, Lemont, IL 60439 USA}
 \affiliation{Northwestern Center for Water Research, Northwestern University, Evanston, IL, 60201}


\date{\today}

\begin{abstract}
In this work we introduce an open-ended question benchmark,
ALDbench, to evaluate the performance of large language models
(LLMs) in materials synthesis, and in particular in the field of atomic layer deposition, a thin film growth technique used in energy applications and microelectronics. Our benchmark comprises questions with a level of difficulty ranging from graduate level to domain expert current with the state of the art in the field. Human experts reviewed the questions along the criteria of difficulty and specificity, and the model responses along four different criteria: overall quality, specificity, relevance, and accuracy. We ran this benchmark on an instance of OpenAI’s GPT-4o. The responses from the model received a composite quality score of 3.7 on a 1 to 5 scale, consistent with a passing grade. However, 36\% of the questions received at least one below average score. An in-depth analysis of the responses identified at least five instances of suspected hallucination. Finally, we observed statistically significant correlations between the difficulty of the question and the quality of the response, the difficulty of the question and the relevance of the response, and the specificity of the question and the accuracy of the response as graded by the human experts. This emphasizes the need to evaluate LLMs across multiple criteria beyond difficulty or accuracy. 
\end{abstract}

\maketitle

\section{\label{sec:intro}Introduction}

The past couple of years have seen a surge of interest in large language models (LLMs) both for both general purpose and scientific applications, including areas such as chemistry and materials science. Views on the usefulness of LLMs in chemistry and materials domains  are somewhat mixed. On one hand, recent papers have framed LLMs as superhuman chemists\cite{chembench} or have posited that the future of chemistry is language.\cite{language} This is motivated by the growing number of works that have explored the use of LLMs for tasks such as materials property prediction, reaction optimization,  or the design of novel materials\cite{inversedesign,matbenchmark,autonomous}, as well as the performance of state of the art LLMs on scientific benchmarks such as ChemBench\cite{chembench}.

This optimistic view is balanced by a more nuanced or skeptical approach towards LLMs, supported by their shortcomings on some of the same tasks. For instance, when analyzing the performance of LLMs in chemistry tasks comprising the ChemBench benchmark, the authors found that the models are still limited in their ability to address knowledge-intensive questions. Likewise, in the same work they questioned whether the model's performance in tasks involving molecular structures based on SMILES was consistent with any chemical reasoning ability\cite{chembench}.

One of the areas still lacking relevant benchmarks is materials synthesis and, in particular, thin film growth. In this work, we introduce a new benchmark to evaluate LLM's knowledge and expertise in the field of atomic layer deposition (ALD), a gas phase thin film growth technique that is based on the self-limited surface reactions of gaseous precursors.\cite{aldreview} Beyond its applied interest in areas such as energy and microelectronics\cite{alvaro}, ALD as a field brings together some key topics that are representative in chemistry-driven synthesis, such as heterogeneous reactions, metalorganic and inorganic molecules, materials microstructure, transport processes, and application-specific knowledge. Consequently, results on the LLM’s proficiency in ALD can provide insights into the model capabilities in these areas. ALD is also a quantitative technique, with well-defined growths per cycle for many different processes. This makes it an interesting model system to explore LLM’s ability to learn specific information about materials synthesis.

In this work we have the following two goals. First, we seek to develop a benchmark that can help us probe the capabilities of LLMs in the context of atomic layer deposition beyond traditional natural language processing or multiple choice question approaches\cite{matsciml}. Here we have focused on an open-ended question format where both the questions and the model responses are graded by human domain experts.

Second, we seek to understand LLMs performance in the context of ALD, focusing primarily on knowledge and research assistance. The open answer question format allows us to select questions that are representative of the potential use of LLMs by any user. We also focused on questions whose answers require either significant domain expertise or would be expected from an expert in an adjacent field. This sets our approach apart from other benchmarks in the literature that are usually focused on general knowledge questions at the undergraduate and graduate levels.

\section{\label{sec:methods}Methodology}

\begin{table*}
\caption{\label{tab:table1}Grading criteria and rubric for the evaluation of the questions in the benchmark and the LLM response}
\begin{ruledtabular}
\begin{tabular}{llll}
\multicolumn{4}{c}{\emph{Criteria and rubric for the questions}} \\
\hline
Difficulty & How hard is the question & 1-Easy, early graduate
& 5-Hard, top expert \\
Specificity & How specific is the question & 1-General &
5-Specific, quantitative \\

\hline
\multicolumn{4}{c}{\emph{Criteria and rubric for the responses}} \\
\hline
Quality & Overall quality of the response & 1-Very low quality
& 5-Excellent \\
Specificity & How specific is the response & 1-Too broad &
5-Targeted \\
Relevance & Is the response relevant & 1-Irrelevant fluff &
5-Relevant answer \\
Accuracy & How accurate is the response & 1-All made up &
5-All correct\\
\end{tabular}
\end{ruledtabular}
\end{table*}

The benchmark was created using human-generated questions: six of the eight co-authors of this work contributed to the curation of the dataset. They are each domain experts in the area of ALD, and were asked to write "questions that a researcher or a graduate student who is not familiar with a specific process/application would ask an AI assistant." Questions were written independently and collated to create an open answer dataset comprising 70 questions. Any modifications to these questions were only to ensure that each prompt asked a single question with a verifiable answer.

We ran this benchmark using an instance of OpenAI's GPT-4o large language model. For our implementation, we leveraged the internal generative AI interface available to researchers within Argonne National Laboratory, called Argo. Private copies of multiple LLMs are available through Argo, offering a data-secure and experimentally controllable environment for research-grade LLM interactions. This custom platform provides a standard API through which we pass the necessary parameters for the LLM inference process. Argo acts as a gateway to a selected LLM by making the appropriate inference call with our submitted content and returns the response from the LLM through the API.

For this experiment, no system prompt was provided, so the LLM responses were intentionally not guided by additional context or direction. A temperature value of 0.1 and a top\_p of 0.9 were assigned for all LLM inference queries, allowing for a consistent configuration to ensure less hallucination and the model to respond with the most likely tokens. Finally, each query from the benchmark was submitted to the LLM independently of all others so that a continuous conversational thread with the LLM was not implemented during the experiment. This approach made certain that the generated answer from any query did not provide inappropriate context or direction for subsequent queries from our benchmark.

Seven of the eight coauthors, all domain experts in the field, evaluated the quality of the open-ended questions by manually reviewing and scoring the answers provided by the LLM. These domain experts were requested to score the answers according to four criteria: overall quality, relevance, specificity, and accuracy, using a Likert scale of 1 to 5. In addition to grading the answers, the domain experts were asked to grade all questions in the dataset along two criteria: difficulty, and specificity of the required response, also using a 1 to 5 Likert scale. A rubric for each of the criteria is shown in Table \ref{tab:table1}. The domain experts performed their reviews independently. 

Composite benchmark scores in each of the criteria were calculated considering the average score of each domain expert. This approach assigns equal weight to the experience of each domain expert with the LLM regardless of the number of responses reviewed. In addition to the statistical analysis of the responses, we subsequently carried out an in-depth analysis focused on hallucination as well as contextual and logical issues with specific answers.

\section{Results}

\subsection{\label{sec:aldbench}Description and analysis of ALDbench}

Through the process described in Section \ref{sec:methods} we compiled 70 open-ended questions. These are shown in Tables \ref{tab:table2}-\ref{tab:table5}. An \emph{a posteriori} analysis of the questions lead us to group these across four categories: 1) "how to grow", where the query pertains to how to grow a specific material using ALD; 2) "specific questions about ALD processes", comprising more detailed questions looking for specific information about a process or materials, including growth per cycle, microstructure, or nucleation behavior; 3) "general ALD knowledge", where questions tend to be broader in nature, ranging from general properties of self-limited processes to questions related to simulations; and 4) "applications",
a smaller subset of questions focused on properties or applications of ALD. The resulting set is not intended to be comprehensive in scope, but instead reflects some of the main areas of expertise of the co-authors that contributed questions to the dataset. It is worth mentioning that it also includes two questions focused on atomic layer etching.

\begin{table*}
\caption{\label{tab:table2}List of the questions in the ALD benchmark for the category "How to Grow"}
\begin{ruledtabular}
\begin{tabular}{lp{17cm}}
& Question  \\
\hline
1 & Can I grow ITO epitaxially by ALD without annealing? \\
2 & Give me a good ALD process for growing strontium titanate by ALD \\
3 & What conditions should I use to grow ZnO epitaxially on sapphire by ALD? \\
4 & Which organic molecules can I use for the molecular layer deposition of organic-inorganic composite materials based on Ti? \\
5 & What is a good thermal atomic layer etching process for copper?\\
6 & What is a good ALD process for MgF2?\\
7 & What is a good atomic layer etching process for cobalt? \\
8 & What precursors can be used to grow barium oxide via ALD? \\
9 & How do I grow MoS2 by ALD? \\
10 & What molybdenum precursors can I use to grow MoS2 via ALD?\\
11 & How many ALD cycles will I need to grow a monolayer of MoS2?\\
12 & Does MoS2 ALD with molybdenum hexacarbonyl as the Mo precursor require a nucleation layer?\\
13 & How can I increase crystallinity in ALD MoS2 without thermal annealing treatment? \\
14 & How do I grow Al2O3 via ALD? \\
15 & How do I grow ZnO by ALD?\\
16 & How do I grow Co3O4 via ALD?\\
17 & What are the precursors I can choose from and the fabrication condition required to grow iridium via ALD? \\
18 & How can I grow metal Ru thin films by ALD?\\
19 & How can I grow metal Co thin films by ALD? \\
20 & What co-reactants are needed to grow Co metal films using ALD? \\
21 & What co-reactants are needed to grow W metal films using ALD?\\
22 & What co-reactant should I use for Pt ALD with Pt(MeCp)Me3?\\
\end{tabular}
\end{ruledtabular}
\end{table*}

\begin{table*}
\caption{\label{tab:table3}List of the questions in the ALD benchmark for the category "Specific ALD process"}
\begin{ruledtabular}
\begin{tabular}{lp{17cm}}
& Question \\
\hline
23 & What is the growth per cycle for the ALD of MgO with bis-cyclopentadienyl magnesium and water at 200 C? \\
24 & What is the out of plane preferential orientation of a ZnO film grown from diethyl zinc and water? \\
25 & What mass gain should I expect in QCM during the ALD of TiO2 from TiCl4 and water?  \\
26 & What values of the sticking coefficient should I use to model the ALD of Al2O3 from TMA and water at atmospheric pressure? \\
27 & What is the typical growth per cycle (GPC) of MoS2 ALD?  \\
28 & What is the temperature window for MoS2 ALD? \\
29 & Do certain ALD precursors give more crystalline MoS2 thin films?  \\
30 & How much nucleation delay is there in MoS2 ALD?  \\
31 & Why can TMA react with H2O, O2 plasma, or O3 to form Al2O3 but shows limited reactivity (or not) with O2?  \\
32 & In DFT calculations, why does the ruthenium precursor, tricarbonyl ($\eta$4-2-methylene-1,3 propanediyl) ruthenium(II), C4H6Ru(CO)3), show favorable adsorption on Ru and Mo surfaces but not on SiO2 or MoO3? \\
33 & How can we estimate the decomposition temperature of tricarbonyl ($\eta$4-2-methylene-1,3 propanediyl) ruthenium(II), C4H6Ru(CO)3)? \\
34 & How can we define the adsorption behavior of tricarbonyl ($\eta$4-2-methylene-1,3-propanediyl) ruthenium(II), C4H6Ru(CO)3) on surfaces through ligand dissociation, and which ligand dissociates first?  \\
35 & Why is ALD of metallic aluminum challenging?  \\
36 & What is the maximum growth temperature for Al2O3 ALD with TMA and H2O?  \\
37 & What temperature should I heat my precursor bubbler to when performing In2O3 ALD with InCp?  \\
38 & What are the strongest peaks observed in FTIR spectra during in situ measurements of ALD Al2O3 using TMA and H2O? \\
\end{tabular}
\end{ruledtabular}
\end{table*}

\begin{table*}
\caption{\label{tab:table4}List of the questions in the ALD benchmark for the category "General"}
\begin{ruledtabular}
\begin{tabular}{lp{17cm}}
& Question \\
\hline
39& Can I get inhomogeneous growths over large substrates using an ALD process that is truly self-limited? \\
40 & How can I model precursor-surface interactions during ALD using DFT? \\
41 & What types of plasma sources are available in ALD? \\
42 & What is a hollow cathode plasma source used in ALD? \\
43 & Which plasma parameters can be controlled in a plasma ALD ? \\
44 & What are the basic differences between CVD and ALD? \\
45 & How to eliminate temperature condensation in an ALD reactor? \\
46 & What is a plasma artifact in ALD process? \\
47 & What materials can be synthesized via ALD?  \\
48 & What materials cannot be synthesized via ALD? \\
49 & What is the typical temperature window for ALD? \\
50 & Does the reactor pressure influence the growth rate during the ALD process? \\
51 & How does the temperature effect the growth of films during the ALD process? \\
52 & What characterization can be done to differentiate between Atomic Layer Deposition and Sequential Infiltration Synthesis process? \\
53 & Can we do ALD under atmospheric pressure? \\
54 & Why is initial growth rate different from steady state growth per cycle in a ALD process? \\
55 & How can we reduce oxidation to maintain a metallic thin film during the metal ALD process? \\
56 & Can co-reactants (sequential pulses) be used to modify the surface during ALD to adjust the growth per cycle? \\
57 & If contamination from precursor fragments remains in the ALD film, how does it affect the crystal structure of the ALD film? \\
58 & How can we remove carbon (C) from precursors in ALD thin films? \\
59 & Can we estimate the surface coverage of a precursor from DFT calculations? \\
60 & How can we distinguish film formation by ALD or CVD at high temperatures?  \\
61 & What features of an ALD process dictate whether the process can be scaled up for commercial production? \\
62 & Why is it that sometimes an ALD process that works well on planar substrates fails when attempting to coat high surface area powders? \\
63 & What controls the maximum aspect ratio that an ALD process can coat conformally? \\
64 & Who invented ALD? \\
65 & What is a major source of signal instability when performing in situ QCM measurements of ALD? \\
\end{tabular}
\end{ruledtabular}
\end{table*}

\begin{table*}
\caption{\label{tab:table5}List of the questions in the ALD benchmark for the category "Applications"}
\begin{ruledtabular}
\begin{tabular}{lp{17cm}}
& Question \\
\hline
66 & What is a good test structure for characterizing the ferroelectric behavior of HfZrO2 thin films? \\
67 & What metal oxide have been deposited via ALD for membrane application?  \\
68 & What is the role of metal oxide deposited via ALD on membranes for water filtration? \\
69 & Has ALD been used for modification of gas separation or solvent separation membranes? \\
70 & What is the typical electron mobility of MoS2 thin films prepared by ALD? \\
\end{tabular}
\end{ruledtabular}
\end{table*}

\begin{figure*}
\includegraphics[width=14cm]{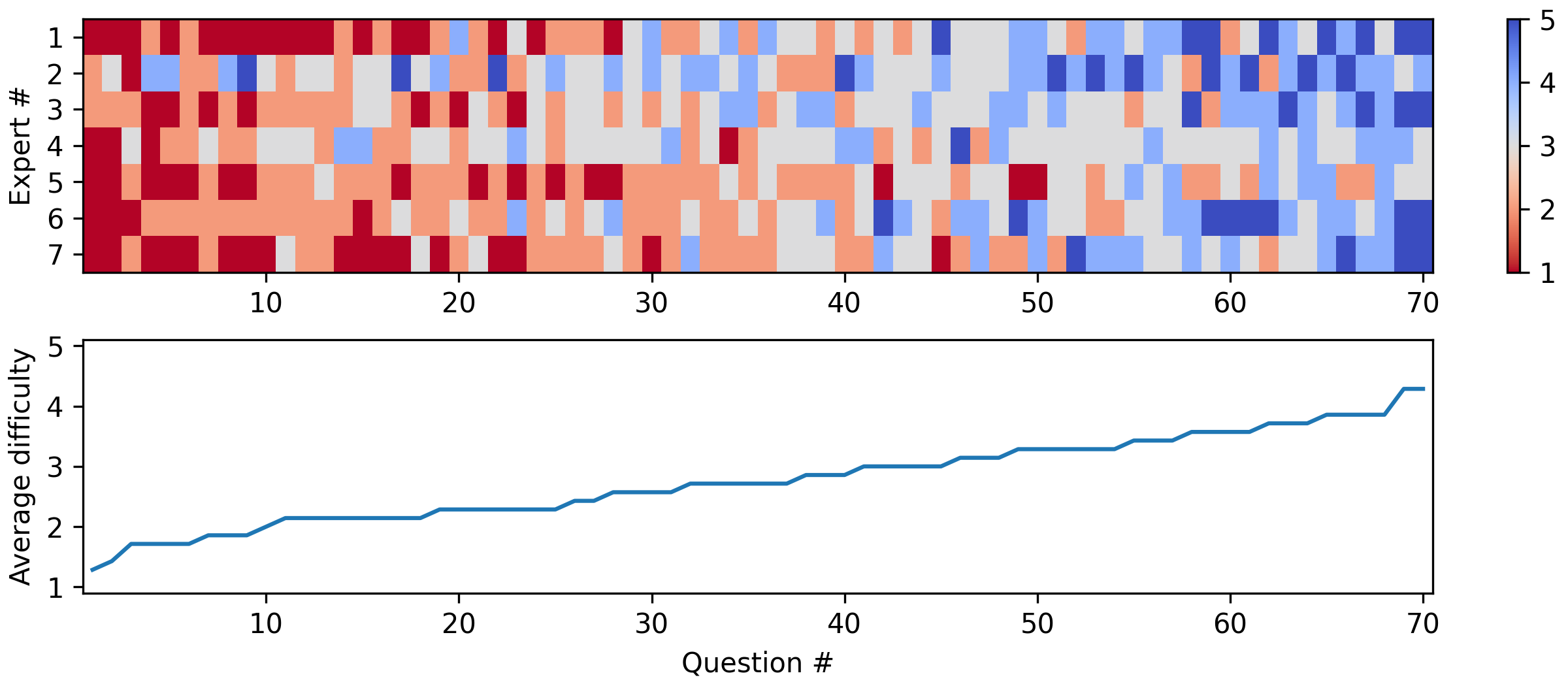}
\caption{\label{fig1}Question grading in order of increasing average difficulty. The questions in the benchmark are well distributed across the 1 to 5 scale in difficulty, with a top rating of 5 corresponding to questions that only a few experts in the field would be able to answer. The color map provides a visualization of the grading by each domain expert, showing some significant dispersion in scores for intermediate values.}
\end{figure*}

The domain experts graded each of these questions according to two different criteria: difficulty and specificity of the question. A key differentiating factor of this benchmark with respect to other approaches is that hard questions require state-of-the-art knowledge  available to only a few experts in the field and require a dedicated search to find them. This expands the range of questions in traditional LLM benchmark beyond the usual ceiling of graduate-level questions. In Figure \ref{fig1}
We show a color map with the difficulty score of each of the domain experts. Questions are presented in increasing order of average difficulty. The corresponding average is also shown in the Figure. The average difficulty for the dataset across all evaluations was 2.7, with the benchmark receiving 37 independent ratings (7.5\%) in the top category. Figure \ref{fig1} also shows a significant dispersion among domain experts. This highlights the breadth of scope that can be found even in highly focused benchmarks such as the one explored in this work, particularly as the difficulty and specificity of the questions increase. 

\begin{figure*}
\includegraphics[width=14cm]{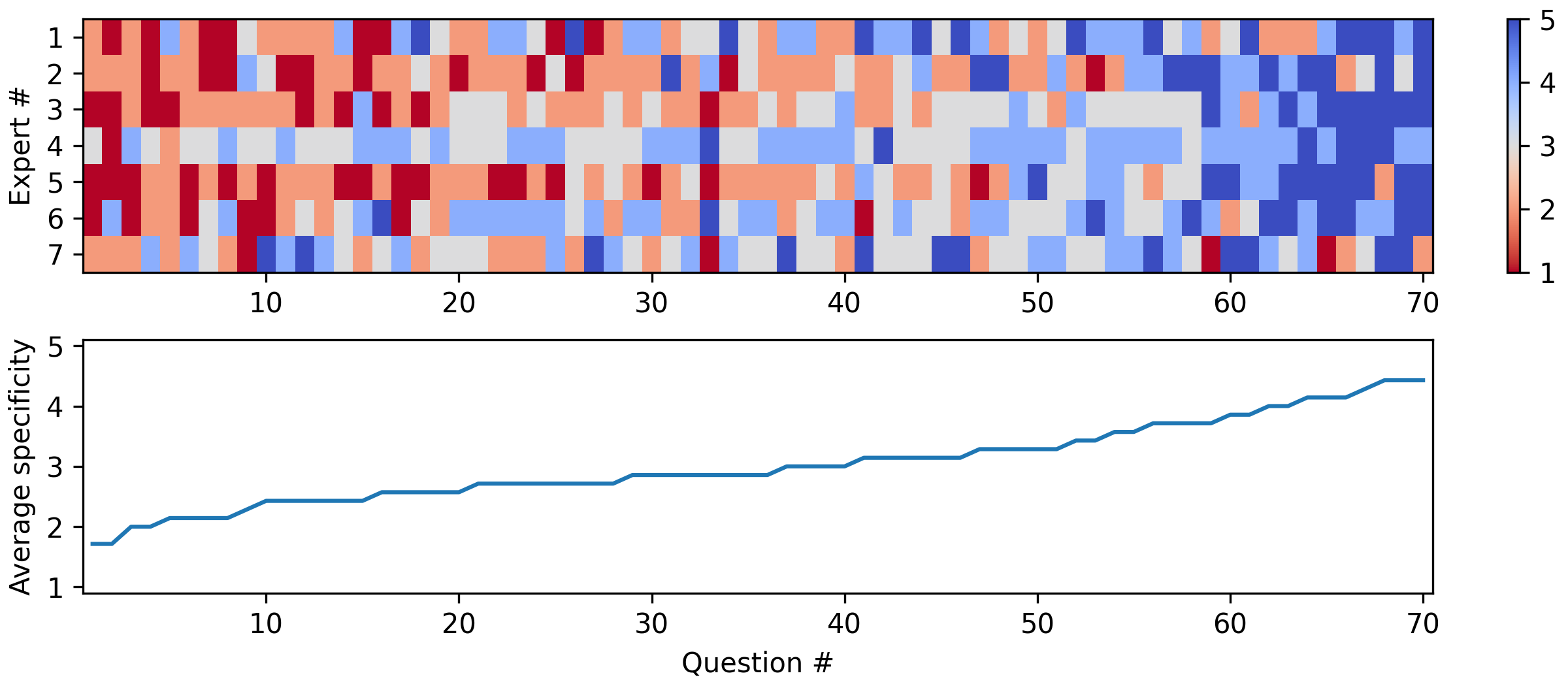}
\caption{\label{fig2}Question grading in order of increasing average specificity. Compared to the difficulty score, questions tend to be more clustered around the intermediate values. The color map provides a visualization of the grading by each domain expert, showing some significant dispersion in the scores.}
\end{figure*}

Similarly, in Figure \ref{fig2} we show the breakdown of the specificity scores by domain expert. The average specificity score is 3.0. Compared to difficulty, the specificity scores tend to be more clustered around intermediate values. 

\begin{figure}
\includegraphics[width=7cm]{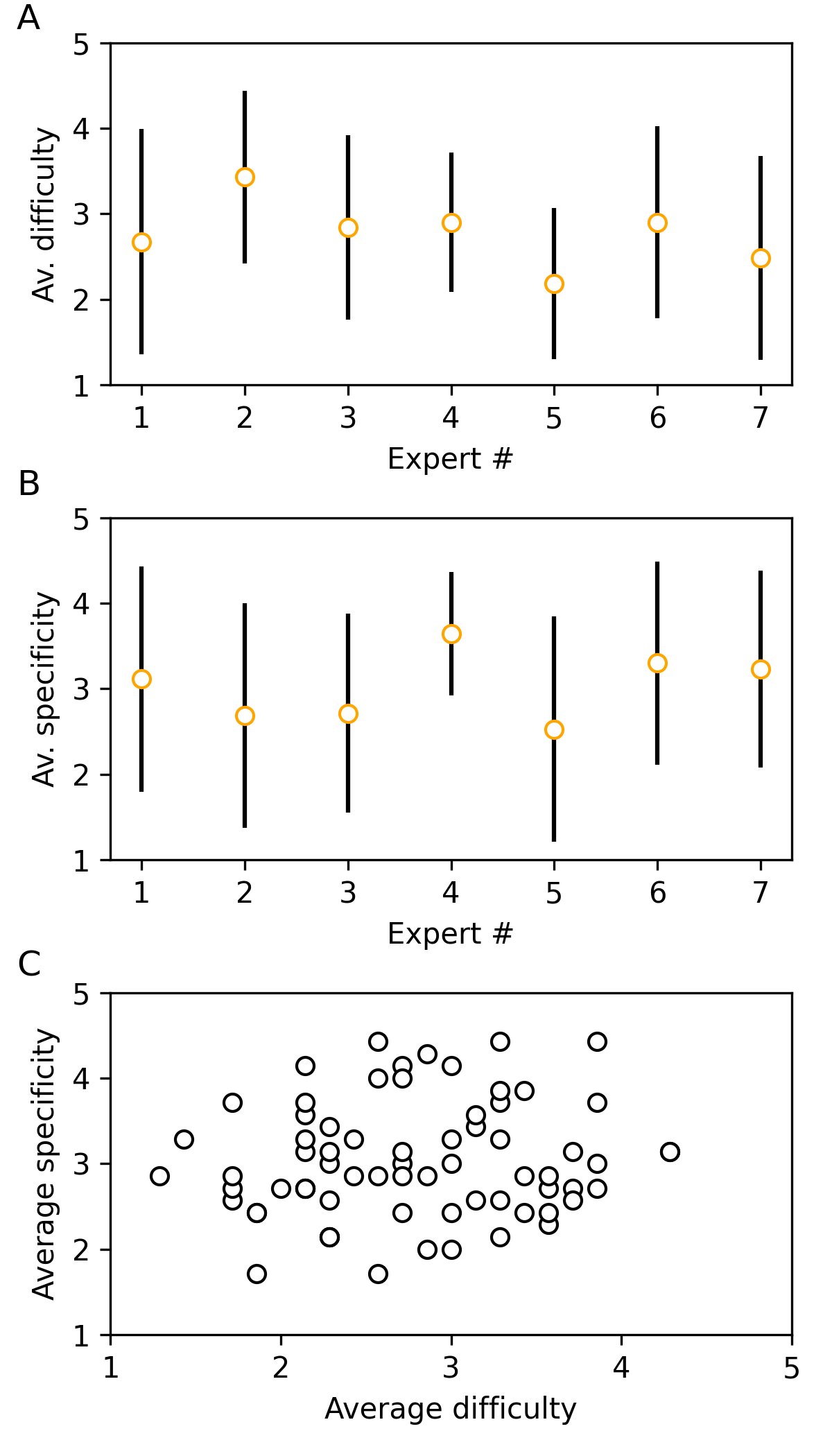}
\caption{\label{fig3}A) Average difficulty score of the benchmark questions assigned by each of the domain experts (B) Average specificity of the question (C) correlation between question difficulty and specificity for all questions in the benchmark. Each value is the average score across all domain experts.}
\end{figure}

Based on the dispersion in Figures \ref{fig1} and \ref{fig2}, we explored how the average score differs across different domain experts. In Figures \ref{fig3}(A) and \ref{fig3}(B) we show the average scores together with errors bars indicating the standard deviation across all the benchmark questions. Despite individual differences observed in Figures \ref{fig1} and \ref{fig2}, the aggregated values are similar across domain experts, with the average values of difficulty and specificity being close to each other. The differences in aggregate tend to be smaller than the standard deviation of the scores.

One motivation for exploring both dimensions is that difficulty and specificity do not have to be correlated: a question may be simple but require requires a very specific answer. In Figure 3(C), we show the correlation between the average difficulty and specificity of each of the questions in the benchmark. The Pearson correlation coefficient is 0.12, indicating a very low correlation between these two values. 

\subsection{\label{sec:quant}Quantitative evaluation of GPT4o performance}

The responses from the LLM were gathered and distributed to all domain experts. Experts were free to review a subset of questions. Through this process, we gathered 236 independent reviews.

Responses were evaluated in terms of four criteria: overall quality, specificity of the answer, relevance of the answer, and accuracy. Our motivation to include specificity and relevance is that these are two criteria that are hard to gauge using autocompletion or multiple choice question benchmarks, and yet they can factor in on the perception of response quality. In particular, specificity refers to the LLM’s ability to provide specific responses to a question vs more generic or discursive answers. With the criterion of relevance, we are addressing the model’s ability to pick the most relevant example or response among many (i.e. the most commonly used ALD process for a specific material).

\begin{figure}
\includegraphics[width=7cm]{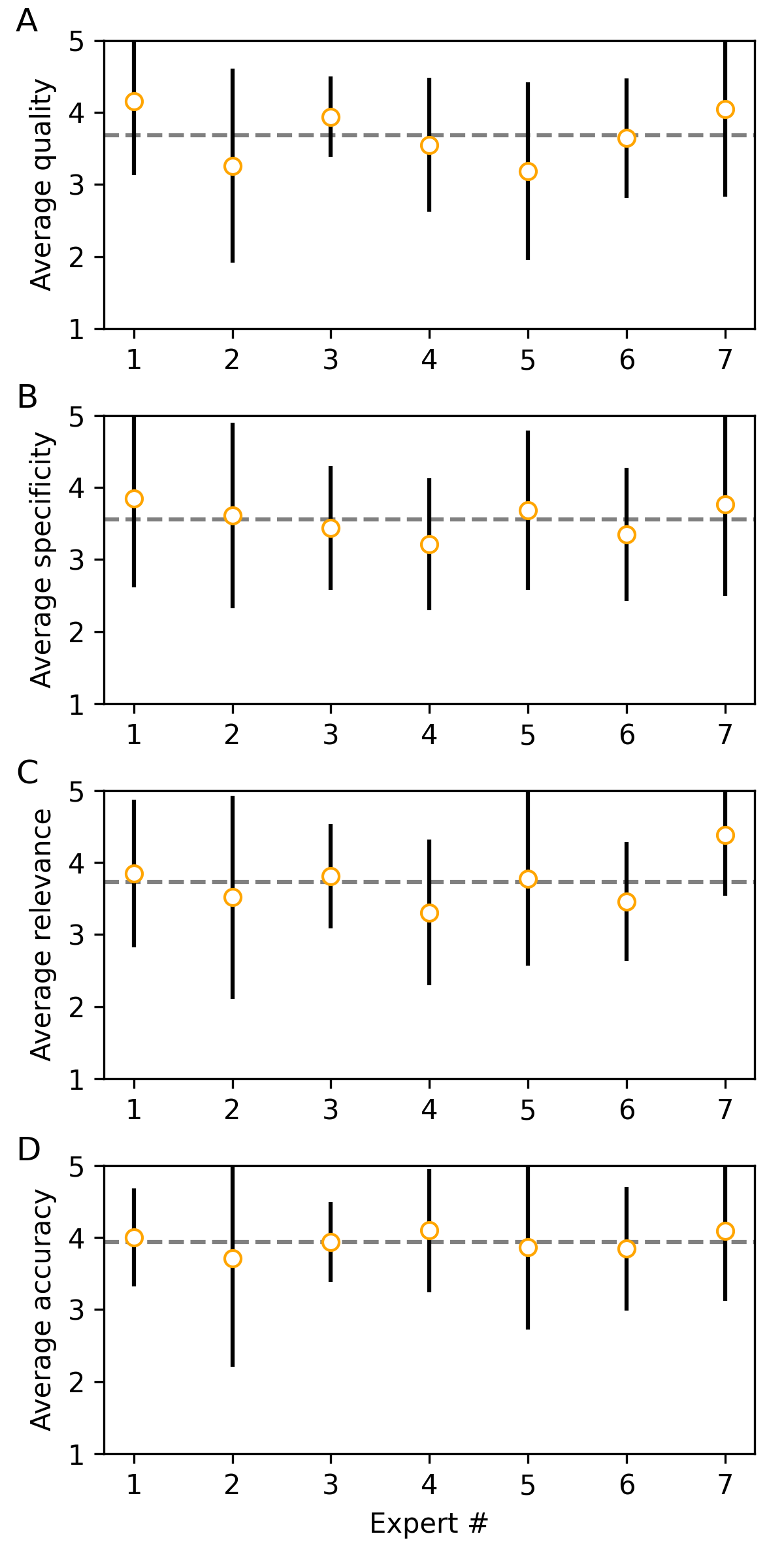}
\caption{\label{fig4}Average scores of GPT4o responses for each of the domain experts: (A) Overall quality (B) Specificity of the response (C) Relevance (D) Accuracy. Error bars represent one standard deviation. Dashed lines represent the aggregated scores of all experts.}
\end{figure}

In Figure \ref{fig4} we show the average scores of the responses generated by our GPT4o instance in each of these four criteria for all our domain experts. The error bars represent the standard deviation. In terms of overall quality, the responses from GPT4o received a composite score of 3.7, with the average score of each domain expert ranging from 3.2 to 4.1. With the 1-5 scale used in this work, this represents an above average grade. 34 scores, or 14\% of the reviews, were below average (either 1 or 2). This resulted in 25 of the questions (36\%) receiving at least one below average grade in terms of GPT4o’s overall quality of the response.

The aggregated average scores in specificity, relevance, and accuracy were 3.6, 3.7, and 3.9 respectively. The breakdown by domain expert of these scores are shown in Figures \ref{fig4}(B)-(D). Interestingly, GPT4o responses were graded higher in accuracy than in specificity or relevance. Agreement among the domain experts was greater for the specificity and accuracy criteria, while the relevance scores showed a variance similar to that of the average quality of the responses. 29 (41\%), 28 (40\%), and 18 (26\%) of the questions received at least one below average grade in specificity, relevance, and accuracy, respectively. 

\begin{figure}
\includegraphics[width=7cm]{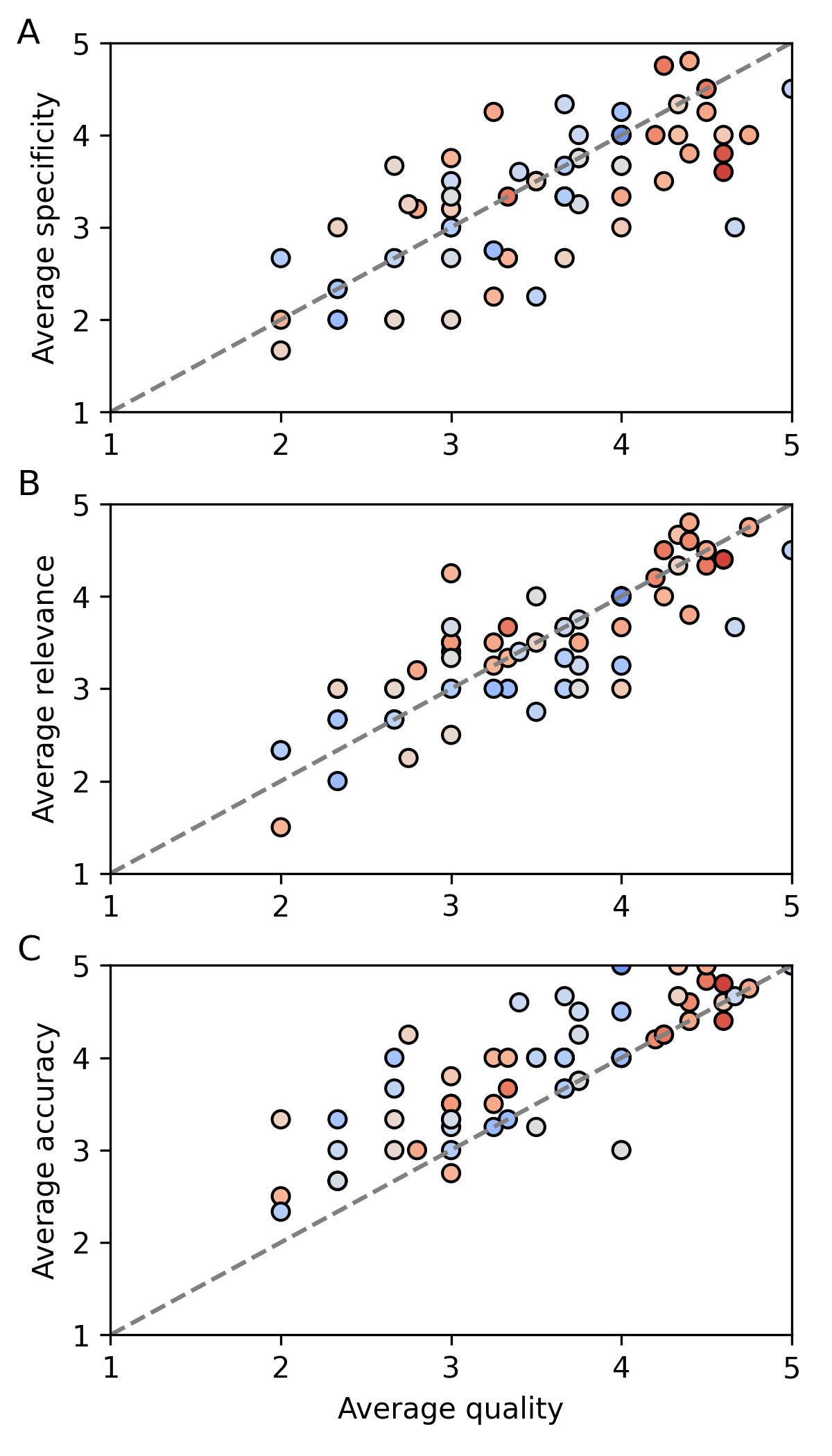}
\caption{\label{fig5}Correlation between the average quality of the GPT4o responses and the other three criteria: (A) Specificity of the response (B) Relevance of the response (C) Accuracy. Points are colored by average difficulty score of each question, using the same color map as in Figures \ref{fig1} and \ref{fig2}
(red - easy, blue - hard). Dashed line represents the $Y=X$ line.}
\end{figure}

To better understand how the specificity, relevance, and accuracy criteria correlate with the overall quality, we calculated the average of the domain expert scores for each of the questions. In Figure \ref{fig5}, we show the correlation between the average quality score of each response and their average specificity, relevance, and accuracy [Figures \ref{fig5}(A)-(C)]. We notice that accuracy metric consistently outperforms the quality score, while the relevance scores tend to align better with the overall quality. The corresponding Pearson correlation coefficients are 0.75 (specificity), 0.84 (relevance) and 0.83 (accuracy).

Finally, we explored whether there was a correlation between the LLM response scores and the difficulty and specificity of each question as scored by each domain expert. We broke down all scores into two groups: one including above average grades (5 and 4) and at or below average grades (3-1). We then built 2x2 contingency tables to explore significant correlations between the scores of the responses and those of the benchmark questions, shown in the appendix. The resulting p-values, obtained from applying the Fisher exact test to each table, are summarized in Table \ref{tab:table6}.

We observed significant differences (p-values < 0.05) in three cases: 
\begin{enumerate}
\item The overall quality of the answers correlated with the difficulty assigned to each question. That is, above average quality scores tend to go to questions rated as easier by the domain experts.

\item We observed significant differences in the relevance score of the responses depending on whether their difficulty score was above average or at or below average: questions rated at or below average in difficulty had a significantly large share of above average relevance scores.

\item We found a strong anti-correlation between the specificity of the question and the accuracy score of the responses. Responses highly rated in accuracy tend to come from questions rated lower in specificity. Differences here were the most significant, with a p-value lower than 0.01.
\end{enumerate}

\begin{table}
\caption{\label{tab:table6}Correlation between the different response grading and question grading criteria considered in this work. Listed p-values are obtained from the Fisher exact test using the contingency tables listed in the Appendix}
\begin{ruledtabular}
\begin{tabular}{llc}
Response criteria & Question criteria & p-value\\
\hline
Quality	& Difficulty & 0.033 \\
Quality & Specificity & 0.339 \\
Specificity & Difficulty & 0.778 \\
Specificity & Specificity & 0.280 \\
Relevance & Difficulty & 0.016 \\ 
Relevance & Specificity & 0.342 \\
Accuracy & Difficulty & 0.350 \\
Accuracy & Specificity & 0.007 \\
\end{tabular}
\end{ruledtabular}
\end{table}

\subsection{\label{sec:depth}In-depth analysis of the LLM answers}

In addition to the quantitative results shown in Section \ref{sec:quant}, we have also carried out an in-depth analysis of some of the LLM responses. Questions in the "how to grow" category (Table \ref{tab:table2}) are particularly useful, because they request information about possible ALD synthesis approaches that can be easily validated against the literature and existing databases. Within these questions, we were able to identify a few examples of hallucination, where the LLM response proposed processes that, to our knowledge, have not been reported in the literature. One example pertains the ALD of MgF$_2$. The LLM response mentions the following about candidates for fluorine sources for ALD (see Supporting Information for the full response): “Hydrogen fluoride-pyridine (HF-pyridine) complex, TiF$_4$, or other fluorine-containing gases like NF$_3$ or SF$_6$ can be used. HF-pyridine is often preferred due to its reactivity and ability to form stable MgF$_2$ films”. While the choice of HF-pyridine or TiF$_4$ are factually correct\cite{hfpyridine,TiF4}, we could not find any precedent of an ALD process for MgF$_2$ utilizing either NF$_3$ or SF$_6$ as co-reactants (See for instance Ref \cite{alddatabase}).
Another example of a factually incorrect ALD process can be found on the LLM’s response about ALD processes for Co$_3$O$_4$. The response includes cobalt(II) nitrate as a precursor. Again, to our knowledge could not find references in the literature to such ALD process. In total, we were able to identify at least five instances of suspected hallucinations in the LLM responses.

We have also observed instances of the model struggling with chemical nomenclature: for instance it failed to recognize "Bis(tert-butylimido) bis(dimethylamido) molybdenum  (Mo(NtBu)$_2$(NMe$_2$)$_2$)" as an alkylamide precursor. In a separate instance, the response reverses the order of the ligands and the metal, such as “(Cp)$_2$Mg”. These are minor concerns compared to factual mistakes, but together these observations highlight the importance of developing benchmarks that prompt LLMs to generate outputs involving chemicals, and particularly metalorganic compounds.

A second recurring pattern in the LLM responses involves too vague or too broad quantitative information. Examples include ranges instead of precise values for the growth per cycle of ZnO ALD from diethyl zinc and water or temperature conditions for an ALD process. These are all factually correct responses, but they fall short of the expected level of precision. These observations are consistent with the significant anticorrelation observed in Section \ref{sec:quant} between the accuracy scores of the responses and the specificity scores of the questions (more specific questions tend to receive lower accuracy scores).
Finally, the responses tend to list precursors and co-reactants as if they are interchangeable. While it is true that many precursors may be reactive towards the same co-reactants (i.e. water and ozone in many oxide ALD processes), this is not generally the case. Also, growths per cycle may differ depending on the co-reactant used, for instance in the case of trimethyl aluminum / water vs trimethylaluminum / ozone\cite{TMAozone}. This structuring of the information is observed in almost all responses involving descriptions of ALD processes (see Supporting Info). On the other hand, the LLM responses do a good overall job emphasizing the key features of self-limited processes, except for isolated instances where the model is asked to reason about the consequences of self-limiting reactions.

\section{Discussion}

\subsection{ALDbench as a domain-specific open-ended question benchmark}

Open-ended question benchmarks such as the one presented in this work are much harder to scale than conventional benchmarks used in automatic evaluation, as they usually require evaluation by domain experts. On the other hand, by exploring domain-specific open-ended questions, we were able to evaluate aspects of LLM’s performance that cannot be evaluated using natural language processing tasks or multiple-choice questions. Through our benchmark we targeted evaluation criteria in the model responses such as relevance or specificity of the response. We were also able to gauge the model’s propensity for hallucinating processes, and the ability to reason about the fundamental aspects of ALD.

An analysis of the scores of the benchmark questions lead us to two key conclusions: first, specificity and difficulty are largely uncorrelated. This highlights the need of considering other criteria beyond difficulty when designing a benchmark. Subsequent analysis of the GPT4o responses showed significant correlations between questions rated as above average in the difficulty and specificity criteria and scores of the model responses.  Second, we observed high variability on the assessment of difficulty by different domain experts. We believe that this is due to the high degree of specialization in many scientific disciplines once one moves beyond general knowledge. This emphasizes the need for incorporating feedback from multiple domain experts even when targeting a well-defined field such as ALD. Based on the distribution of the scores of the benchmark questions, the 70 human-generated questions resulted in a well-balanced set.

\subsection{GPT4o as a baseline LLM for ALD and materials synthesis}

We used GPT4o as our baseline LLM for our benchmark. This model did not have access to external search or data, so the responses reflect the model’s intrinsic ability to generate text based on its training. On the one hand, the fact that the model received an aggregated passing score in all four evaluation criteria is certainly a technical achievement. On the other hand, between 15\% and 35\% of the questions received at least one below-average score in one of the evaluation criteria. The model also hallucinated some responses, particularly those involving chemical precursors. Some hallucinations are consistent with the model’s inability to identify the proper context. For instance, the mention of NF$_3$ and SF$_6$ as possible fluorine sources for the ALD of metal fluorides makes sense given that SF$_6$ plasmas have been used
for ALD of AlF$_3$ and LiF\cite{AlF3,SF6}. NF$_3$ also appears in the literature in the context of etching of ALD films\cite{NF3etching}. Thus, there is a real association between both compounds and ALD that gets misinterpreted during the response generation process. This highlights one of the challenges of existing LLMs, particularly those  not specifically trained for scientific applications, when dealing with complex scientific problems. This is consistent with prior observations on chemistry benchmarks\cite{chembench}.

Finally, we must point out that there are strategies that can further improve the model's performance that we did not explore in this work. One approach is prompt engineering, wherein queries can be prefaced by text that conditions the answer the model provides (i.e. "you are an awesome model and one of the leading experts in ALD in the world, please answer the following question to the best of your abilities:"). A second method is through fine-tuning the hyperparameters used as inputs for the generative process. A third approach is augmenting LLMs with tools that provide external knowledge\cite{augmented}. One final factor that we haven’t considered is the stochasticity in text generation: the sequence of tokens is generated probabilistically, thus potentially leading to different responses to the same query in each interaction. We are currently in the process of generating a dataset to evaluate the prevalence and reproducibility of fabricated outputs using these strategies. The results will be presented in a future work.

\section{Conclusions}

In this work we introduced a new open-ended question benchmark for materials synthesis, ALDbench, specifically related to atomic layer deposition. Our benchmark complements other approaches such as MatSciML\cite{matsciml}, which are more geared towards natural language processing tasks. Instead, our benchmark focuses primarily on knowledge-based questions that experts would likely ask an AI expert. While this does not cover other potential uses of LLMs such as the design of new processes or the property prediction currently being pursued in the literature, we believe that a strong knowledge-based foundation is critical for the success of these more advanced applications. Consequently, benchmarks such as ALDbench will likely play an increasingly important role when evaluating the potential of LLMs for different scientific domains.

\begin{acknowledgments}
This research is based upon work supported by  Laboratory Directed Research and Development (LDRD) funding from Argonne National Laboratory, provided by the Director, Office of Science, of the U.S. Department of Energy under Contract No. DE-AC02-06CH11357.
The analysis was funded in part by the Advanced Materials for Energy-Water Systems (AMEWS) Center, an Energy Frontier Research Center funded by the U.S. Department of Energy, Office of Science, Basic Energy Sciences.

\end{acknowledgments}

\section*{Author Declarations}

\subsection*{Conflict of interest}

The authors have no conflicts to disclose

\section*{Data Availability Statement}

The data that support the findings of
this study are included in the Supporting Information.

\section*{References}

\bibliography{aldbench}

\appendix

\section{Contingency tables for the question and response criteria}

Here we present the 2x2 contingency tables to explore the
correlation between the question scores, and the score of the
model response across the four different criteria. In order
to calculate these contingency tables, we split all reviews
into two categories: above average, comprising reviews
with a score of 4 or 5, and at or below average, comprising 
the 1 to 3 range.

In each contingency table, we list all the reviews belonging
to each category. This allows us to apply the standard Fisher
exact test to determine if the differences observed in the
way the reviews are distributed among the four categories are
statistically significant.

Table \ref{table:diff} we show the correlations between the difficulty
of the question and each of the four evaluation criteria for the model
responses: quality, specificity, relevance, and accuracy. In Table \ref{table:spec} we show the contingency tables for the question specificity.

\begin{table}[h!]
\caption{\label{table:diff} Contingency table exploring the
correlation between question difficulty and the difference
response evaluation criteria considered in this work}
\begin{ruledtabular}
\begin{tabular}{l|cc}
 & \multicolumn{2}{c}{Question difficulty} \\
Response quality & Above average & At or below average  \\
\hline
Above average & 60 &  38 \\ 
At or below average & 132 & 35 \\
\hline
\hline
 & \multicolumn{2}{c}{Question difficulty} \\
Response specificity & Above average & At or below average  \\
\hline
Above average & 80 &  38 \\ 
At or below average & 82 & 35 \\
\hline
\hline
 & \multicolumn{2}{c}{Question difficulty} \\
Response relevance & Above average & At or below average  \\
\hline
Above average & 61 & 40 \\ 
At or below average & 101 & 33 \\
\hline
\hline
 & \multicolumn{2}{c}{Question difficulty} \\
Response accuracy & Above average & At or below average  \\
\hline
Above average & 43 &  24 \\ 
At or below average & 119 & 49 \\
\end{tabular}
\end{ruledtabular}
\end{table}

\begin{table}
\caption{\label{table:spec} Contingency table exploring the
correlation between question specificity and the difference
response evaluation criteria considered in this work}
\begin{ruledtabular}
\begin{tabular}{l|cc}
 & \multicolumn{2}{c}{Question specificity} \\
Response quality & Above average & At or below average  \\
\hline
Above average & 58 &  40 \\ 
At or below average & 90 & 47 \\
\hline
\hline
 & \multicolumn{2}{c}{Question specificity} \\
Response specificity & Above average & At or below average  \\
\hline
Above average & 70 &  48 \\
At or below average & 78 & 39 \\
\hline
\hline
 & \multicolumn{2}{c}{Question specificity} \\
Response relevance & Above average & At or below average  \\
\hline
Above average & 60 & 41 \\
At or below average & 88 & 46 \\
\hline
\hline
 & \multicolumn{2}{c}{Question specificity} \\
Response accuracy & Above average & At or below average  \\
\hline
Above average & 33 &  34 \\
At or below average & 115 & 53 \\
\end{tabular}
\end{ruledtabular}
\end{table}

\end{document}